\documentclass[sigconf, nonacm]{acmart}

\AtBeginDocument{%
  \providecommand\BibTeX{{%
    \normalfont B\kern-0.5em{\scshape i\kern-0.25em b}\kern-0.8em\TeX}}}

\usepackage{comment}
\usepackage{color}
\usepackage{bm}
\usepackage{array}
\usepackage{booktabs}
\usepackage{multirow}
\usepackage{mathrsfs}
\usepackage{lipsum}
\usepackage{stackengine}

\definecolor{mygray}{gray}{.9}
\definecolor{mygray1}{gray}{.7}
\usepackage{colortbl}
\usepackage{color}

\makeatletter
\newcommand{\thickhline}{%
	\noalign {\ifnum 0=`}\fi \hrule height 1pt
	\futurelet \reserved@a \@xhline
}
\makeatother
\setcopyright{acmlicensed}
\copyrightyear{2018}
\acmYear{2018}
\acmDOI{XXXXXXX.XXXXXXX}

\acmConference[MM'24]{Make sure to enter the correct
  conference title from your rights confirmation email}{October 28 - November 1,
  2024}{Melbourne, Australia.}
\acmISBN{978-1-4503-XXXX-X/18/06}

\begin{document}

\title{Neural Interaction Energy for Multi-Agent Trajectory Prediction}


\author{Kaixin Shen, Ruijie Quan, Linchao Zhu, Jun Xiao, Yi Yang \\
Zhejiang University}



\begin{abstract}
  Maintaining temporal stability is crucial in multi-agent trajectory prediction. Insufficient regularization to uphold this stability often results in fluctuations in kinematic states, leading to inconsistent predictions and the amplification of errors. In this study, we introduce a framework called \textbf{M}ulti-\textbf{A}gent \textbf{T}rajectory prediction via neural interaction \textbf{E}nergy (MATE). This framework assesses the interactive motion of agents by employing neural interaction energy, which captures the dynamics of interactions and illustrates their influence on the future trajectories of agents. To bolster temporal stability, we introduce two constraints: inter-agent interaction constraint and intra-agent motion constraint. These constraints work together to ensure temporal stability at both the system and agent levels, effectively mitigating prediction fluctuations inherent in multi-agent systems. Comparative evaluations against previous methods on four diverse datasets highlight the superior prediction accuracy and generalization capabilities of our model. 
\end{abstract}






\maketitle
\section{Introduction}

Multi-agent trajectory prediction is a fundamental research task with many applications such as autonomous driving~\cite{levinson2011towards}, interactive robotics~\cite{kanda2002development}, particle simulation~\cite{li2020evolvegraph}, and team sports~\cite{hauri2021multi}. It refers to predicting the future positions of multiple agents in a dynamic environment simultaneously, based on their historical positions and agent interactions. This task is critical in many real-world scenarios where a group of agents interacts with each other, giving rise to complicated behavior patterns at the level of both individuals and the whole system. For this reason, it is more challenging to simultaneously predict the future positions of multiple agents compared to single-agent trajectory prediction~\cite{GRIN}.

One of the major obstacles in multi-agent trajectory prediction lies in the frequent fluctuations of kinematic states across consecutive time intervals. These fluctuations pose a significant challenge as they can lead to inconsistencies in the trajectory predictions for each agent. The variability in kinematic states, such as sudden changes in position or velocity, further causes error amplification in sequential predictions~\cite{sun2022IMMA,alahi2016social,yue2022human}. Such fluctuations can be mainly attributed to the absence of proper regularization to temporal dynamics in multi-agent systems.

Motivated by this, we aim to regularize the multi-agent trajectory prediction systems to be temporally stable, enhancing the prediction robustness across diverse scenarios.
Temporal stability refers to the property that the system's kinematic states do not change abruptly or inconsistently over time, but rather approach a steady-state distribution~\cite{chli2003stability}. 
Temporal stability offers two advantages for trajectory prediction in multi-agent systems: 1) Improved prediction accuracy. By enforcing temporal stability in the multi-agent systems, the prediction results would be more resilient to observation noises and error propagation.
2) Enhanced adaptability to diverse environments. 
The presence of temporal stability in multi-agent systems results in more uniform and coordinated movement patterns, enabling the system to adapt seamlessly to a variety of environmental conditions.

In this paper, we propose a framework named MATE for multi-agent trajectory prediction. 
We enhance the temporal stability of the multi-agent system
with an inter-agent constraint and an intra-agent constraint. These two constraints help to preserve temporal stability at the system level and agent level, respectively.



First, we introduce the \textbf{inter-agent interaction constraint} to enhance the stability of agent interactions at the system level. We represent the interactions between agents as the changes of their \textbf{neural interaction energy}, while the system-level dynamics can be considered as the aggregation of neural interaction energy from all agents.
Specifically, the neural interaction energy quantifies agents' motion interactiveness at each step. It also determines the action of agents in future steps. The agents with low neural interaction energy may exhibit reduced movement speed, whereas those with high energy are prone to be more dynamic.
The inter-agent interaction constraint minimizes the change of system-level neural interaction energy, stabilizing the multi-agent movement patterns and ensuring a more coherent evolution of the multi-agent system over time.





Second, we introduce the \textbf{intra-agent motion constraint} to enforce the agent-level stability. 
For each agent, we leverage a temporal motion variance term to measure the consistency of their movement patterns over time.
We derive an approximated motion based on the agent's past kinematic states and its interactions with surrounding agents.
The temporal motion variance term quantifies the discrepancy between the approximated motion and the predicted motion, which helps to identify the agent movement inconsistency. By minimizing this variance, it ensures that the predicted motion aligns closely with the approximated motion, restricting the kinematic states of agents to be coherent and may vary within a plausible range.






Our model outperforms the state-of-the-art methods in multi-agent trajectory prediction across four datasets, i.e., PHASE~\cite{netanyahu2021phase}, Socialnav~\cite{chen2019crowd}, Charged~\cite{GRIN}, and NBA datasets. We further validate the advantages of our method in zero-shot generalization to unseen scenarios.
In summary, the main contributions of this paper are threefold:  
\begin{itemize}
    \item We propose to use neural interaction energy to model the interactions between agents, providing an effective framework to capture the complexities of agent interactions.
    \item We establish an inter-agent interaction constraint and an intra-agent motion constraint, which well preserve the temporal stability at the system level and agent level.
    \item  
    Extensive experiments demonstrate that our model accurately predicts future trajectories and improves the generalization abilities in unseen scenarios.
\end{itemize}

\section{Related Work}
\textbf{Trajectory Prediction.} Trajectory prediction approaches fall into model-based~\cite{helbing1995social} and model-free methods~\cite{tacchetti2018relational,li2020evolvegraph,webb2019factorised,girase2021loki,cao2020spectral,sanchez2020learning,alahi2016social,yue2022human,manh2018scene,giuliari2021transformer,fassmeyer2022semi,GRIN,gupta2018social,kingma2013auto,sohn2015learning,higgins2016beta,quan2021holistic,tang2023collaborative,mao2023leapfrog}. Before the deep learning era, previous works employ various physical models to conduct trajectory prediction. Social force~\cite{helbing1995social} and energy~\cite{karamouzas2014universal} are commonly used to model pedestrian motion. Some works~\cite{van2011reciprocal} exploit the reciprocal velocity obstacles (RVO) model for trajectory prediction. ~\cite{vemula2018social} predicts transitions in pedestrians by applying spatiotemporal graphs, where nodes and edges are represented by RNNs. ~\cite{pettre2009experiment} computes joint motion predictions based on the time of possible collision between pairs of agents. ~\cite{zhao2019multi} propose the Multi-Agent Tensor Fusion encoding, which fuses contextual images of the environment with sequential trajectories of agents. We will include the above discussion and references in the revision.  
The rise of deep learning~\cite{rabinowitz2018machine,deo2018convolutional,chen2019crowd,paszke2017automatic,gupta2018social} has introduced model-free methods into this field. Such a method shifts the focus of modeling agents' trajectories to fitting the distribution of data~\cite{hauri2021multi,zhan2018generating,girase2021loki,huang2019stgat,mohamed2020social}, which improves computational efficiency while reducing the model interpretability. 
GAT-LSTM~\cite{velivckovic2017graph} employs graph attention layers along with LSTMs for prediction. NRI~\cite{kipf2018NRI} designs a VAE model with recurrent GNN networks to encode and predict trajectories. EvolveGraph~\cite{li2020evolvegraph} and fNRI~\cite{webb2019factorised} expand the framework of NRI and introduce additional modules to improve performance. RFM~\cite{tacchetti2018relational} uses a recurrent GNN model with a supervised loss to realize trajectory prediction. IMMA~\cite{sun2022IMMA} introduces a multiplex graph and proposes a progressive layer training strategy. GRIN~\cite{GRIN} employs generative models to learn the distribution of trajectories.
Inspired by traditional dynamics, we combine a novel concept named neural interaction energy and sequential models to predict the changes in agents' trajectories and introduce an inter-agent interaction constraint and an intra-agent motion constraint for preserving temporal stability.

\noindent\textbf{Differential Equations Informed Neural Network.} Recently the combination of deep learning and differentiable equations has earned significant interest. Related research~\cite{karniadakis2021physics,sankaranimpact} can be categorized into different subfields such as deep learning assisted DE~\cite{desmond2019symplectic}, differentiable physics, neural differential equations~\cite{chen2018neural}, and physics-informed neural networks (PINNs)~\cite{raissi2019physics,li2021physics,zubov2021neuralpde}. PINNs are based on the physical prior that constrains the output by calculating the means of a partial differential equation (PDE) system with the assistance of a neural network. Inspired by such research, we propose novel neural differential equations for modeling interactions and establishing motion constraints for agents to predict coherent future trajectories.

\section{Methodology}
\label{Methodology}

\begin{figure*}
  \centering
  \includegraphics[width=1.0\textwidth]{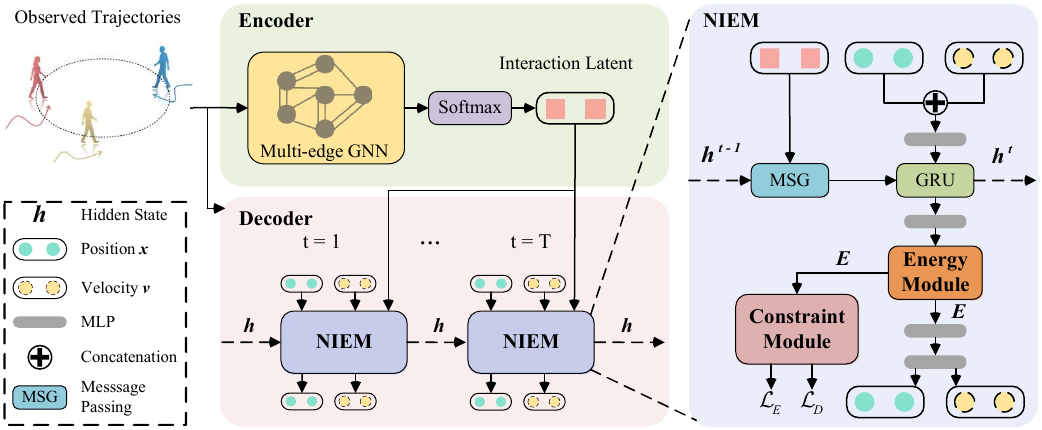}
  \caption{The MATE model consists of (i) a MATE-Encoder responsible for extracting interaction latent between agents, and (ii) a MATE-Decoder integrated with a Neural Interaction Energy Module (NIEM). The NIEM incorporates an Energy Module that extracts the neural interaction energy features $\bm{E}$ of agents. These features are used for the subsequent networks. They are also leveraged to calculate the inter-agent interaction and intra-agent motion constraints through a Constraint Module. Refer to the Appendix for more details about the NIEM.}
  \label{Fig.network}
\end{figure*}

\subsection{Problem Formulation.}

Given an observed trajectory \(\mathbf{X}_i^T = \{\bm{x}_i^1, \bm{x}_i^2, ..., \bm{x}_i^{T_{obs}}\}\) of an agent $i \in N$ across period $T_{obs}$, our goal is to predict the future trajectories of all agents simultaneously. $\bm{x}_i^t \in \mathbb{R}^2$ is the 2D coordinate position of the agent $i$ at time step $t$.
We denote the future trajectory across $T_{pred}$ steps as \(\mathbf{Y}_i^T = \{\bm{x}_i^{T_{obs}+1}, \bm{x}_i^{T_{obs}+2}, ..., \bm{x}_i^T\}\), where $T=T_{obs}+T_{pred}$. Note that for each agent, we only predict a single future trajectory across $T_{pred}$ steps.
In this paper, kinematic states include position and velocity. At each step, the position $\bm{x}$ of an agent indicates its location in the 2D coordinate system. In addition, to represent movement speed, we calculate the velocity $\bm{v}$ of agents at each step based on their 
position and time interval $\Delta t$.




\subsection{Inter-agent Interaction Modeling}

Predicting future trajectories in multi-agent systems necessitates an accurate modeling of agent interactions. Existing approaches often rely on graph neural networks (GNNs) to depict these interactions as latent features of graph edges. However, this representation ignores ensuring the stability of agent interactions. Also, it does not focus on promising a temporally coherent evolution of the system. 
To address these overlooked aspects, we employ a neural interaction energy mechanism to model the interactions of agents, further preserving the temporal stability at the system level.

\noindent\textbf{Neural Interaction Energy $E$.} 
We first present the neural interaction energy to quantify the agent's motion interactiveness, thereby capturing the dynamics of multi-agent interactions and illustrating
how interactions influence agents' future trajectories. 


Neural interaction energy quantifies the motion interactiveness at each step and affects the agent movement in future steps. This motion interactiveness is a function concerning both an agent's kinematic states (position and velocity) and its interaction representation. 
For clarity, we use $E_{i}^{t}$ to denote the neural interaction energy scalar for agent $i$ at step $t$ and use $\bm{E}_{i}^{t}$ to denote the corresponding neural interaction energy features in the neural networks.

Intuitively, the reduction of neural interaction energy will decrease the agent interactiveness, while agents with high neural interaction energy are more likely to move.
In a multi-agent system, the motion of the target agent is affected by other agents. The change in neural interaction energy, reflecting each agent's interactions with others, acts as a measure of the dynamic exchanges within the system. Moreover, the overall dynamics of the multi-agent system can be viewed as the aggregation of the neural interaction energies of all participating agents, providing a comprehensive view of the system's interactive behavior.

\noindent\textbf{Inter-agent Interaction Constraint.}
We introduce an inter-agent interaction constraint to preserve system-level temporal stability based on the interaction between agents. Considering a multi-agent system, agents are moving concurrently and their behaviors are mutually influenced through complex social interactions.
The social interactions among agents indicate the transfer of neural interaction energy, accelerating or decelerating agents. However, due to the complexity of interactions between agents in systems, the neural interaction energy transfer process becomes difficult to analyze from the perspective of the agent level. 

Nevertheless, we model the interactions at the system level. We aim to preserve system-level temporal stability. It means the changes in its neural interaction energy maintain temporal continuity. The target formulation for system-level temporal stability is,
\begin{equation}
     E_{i}^{t+\Delta t} - E_{i}^{t}\rightarrow 0, \quad \text{when } \Delta t \rightarrow 0. 
 \end{equation}
Therefore, over a short time interval, the aggregated neural interaction energy of the system should experience negligible changes, resulting in a nearly constant sum of neural interaction energy changes of agents. That is,
 \begin{equation}
     \Delta \sum_{i=1}^{N}E_{i} = \sum_{i=1}^{N}E_{i}^{t+\Delta t} - \sum_{i=1}^{N}E_{i}^{t}\rightarrow 0, \quad \text{when } \Delta t \rightarrow 0. 
 \end{equation}
Note that the magnitude of change in neural interaction energy differs among different types of agents, necessitating the assignment of weights to achieve normalization. 
Considering that different types of agents have different ranges of positional changes within the same time interval, we employ $\Delta \bm{x}$ as the normalization weight. We have
\begin{equation}
    \label{concrete}
    \sum_{i=1}^{N}\frac{\Delta E_{i}}{\Delta \bm{x_i}} \rightarrow 0, \quad \text{when } \Delta t \rightarrow 0.
\end{equation}
$\Delta E_i$ and $\Delta \bm{x_i}$ denote the change in neural interaction energy and the change in position of agent $i$, respectively.


Consequently, we can model the interactions between agents while preserving the temporal stability at the system level. Formally, the system would have minimal neural interaction energy change in its aggregated interactions of all agents over a short period. Since the time interval is short, $\Delta \bm{x}$ tends to be zero. Not that we use the derivative of the neural interaction energy features concerning the position $\nabla_{\bm{x}}\bm{E}_{i}(\bm{x})$ in implementation to replace the expression in Eq. (\ref{concrete}). That is 
\begin{equation}
\label{inter-agent pde}
\sum_{i=1}^{N} \nabla_{\bm{x}}\bm{E}_{i}(\bm{x}) =  \sum_{i=1}^{N} \lim_{\Delta x \to 0} \frac{\Delta \bm{E}_{i}}{\Delta \bm{x}_{i}} \rightarrow 0.
\end{equation}
We omit the superscript $t$ for simplification. 

In summary, we leverage Eq. (\ref{inter-agent pde}) to ensure the temporal stability of a multi-agent system by minimizing the aggregated changes in neural interaction energy over a short period.
Therefore, by incorporating this inter-agent interaction constraint into model optimization, we preserve system-level temporal stability.
We add this constraint in the form like
\begin{equation}
\mathcal{L}_{E} = {\frac{1}{N}\frac{1}{T}\sum_{i=1}^{N}\sum_{t=1}^{T}\|\nabla_{\bm{x}}\bm{E}_{i}^{t}(\bm{x})\|_2}.
\end{equation}

\subsection{Intra-agent Motion Modeling}

This section introduces to improve intra-agent motion coherence and avoid catastrophic erroneous predictions, preserving agent-level temporal stability.
If the value of the target agent's kinematic states deviates from the normal range (e.g., a pedestrian moves at the speed of a car), it indicates unstable changes and a potential prediction error. We aim to mitigate the prediction discrepancy at each step.
Technically, we define a temporal motion variance term and regularize this variance to be coherent.
This helps to restrict the value of the agent's kinematic states to vary within a plausible range.




The kinematic states of the target agent in the current time step ($\bm{x}^{t}$, $\bm{v}^{t}$) can be determined by
the kinematic states from the previous time step
and the neural interaction energy.
We calculate the derivative of the predicted position $\bm{x}^{t}$ and the change in neural interaction energy $\nabla_{\bm{x}}\bm{E}^{t}(\bm{x})$ concerning the previous step velocity $\bm{v}^{t-1}$. We formulate the following temporal motion variance $u$ for the target agent,
\begin{equation}
\label{intra constraint}
u^t=\gamma \cdot \frac{\partial \nabla_{\bm{x}}\bm{E}^{t}(\bm{x})}{\partial{\bm{v}^{t-1}}} +  \beta \cdot \frac{\partial\bm{x}^{t}}{\partial{\bm{v}^{t-1}}} + \alpha ,
\end{equation}
where $\gamma$ and $\beta$ are scaling scalars and $\alpha$ is the bias term representing a constant correlation between the position and velocity of agents. For clarity, we omit the subscripts $i$ of the variables as agents share the same formula.  
Within a multi-agent environment, an agent's position is not solely determined by its own intention but is also influenced by the interactions with other agents. Consequently, in addition to $\frac{\partial\bm{x}^{t}}{\partial{\bm{v}^{t-1}}} $, our motion variance $u$ introduces an additional term to encapsulate external influences on the agent. Hyperparameters $\gamma$ and $\beta$ are introduced to weigh the significance of these two terms, catering to scenarios where agent behaviors vary based on differing extents of external influences. This refined approach renders our proposed constraints more adaptable for trajectory prediction within multi-agent systems.


In the temporal motion variance $u$, we use the derivative of the predicted output ($\nabla_{\bm{x}}\bm{E}^{t}(\bm{x})$ and $\bm{x}^t$) concerning $\bm{v}^{t-1}$ to measure the prediction's sensitivity.
A high sensitivity means that a minor miscalculation in estimating $\bm{v}^{t-1}$ can magnify into a prominent prediction error. The detailed mathematical derivation of Eq. (\ref{intra constraint}) and determination of hyperparameters $\gamma$, $\beta$, and $\alpha$ can be found in the Appendix. 

By minimizing the temporal motion variance $u$:
\begin{equation}
    \mathcal{L}_{D} =\frac{1}{N}\frac{1}{T}\sum_{i=1}^{N}\sum_{t=1}^{T}\|u_i^t\|_2,
\end{equation}
we reduce the prediction sensitivity and penalize inaccurate or incoherent predictions, achieving agent-level temporal stability.
We add this intra-agent motion constraint to model optimization.

\subsection{Model Architecture}


\textbf{MATE-Encoder with Multi-Edge Graph.} Previous research employs a single-edge GNN to represent interactions between agents. In such a framework, the interaction latent $\bm{z}$ produced by the GNN is normalized and then utilized as a one-hot vector to symbolize interactions for proceeding prediction. However, multiple types of interactions frequently coexist between agents in real scenarios, and a simple one-hot vector fails to encapsulate the intricate properties of these interactions. For instance, in social event systems, factors like familiarity, status, and identities concurrently influence agents' interactions with others.

Addressing this, as shown in Fig.~\ref{Fig.network}, we present a multi-edge GNN, where multiple edges are assigned between agents. Each edge symbolizes a distinct type of interaction and is separately encoded through a unique Multi-Layer Perceptron (MLP) layer. We hypothesize that all types of interactions influence the agents' neural interaction energy and their future trajectories. This fosters a richer representation of the interactions in a given multi-agent system.

Mathematically, given the observation of all agents $\bm{X}^{1:T_{obs}}$, we employ a GNN model to learn the interaction latent as $\bm{z}_{ij} = GNN(\bm{X}_i, \bm{X}_j, \theta)$. Here $\bm{X}_i$ and $\bm{X}_j$ represent the observed trajectory of agent $i$ and $j$ respectively, $\theta$ denotes model parameters, and $\bm{z}_{ij}$ represents the interaction latent between agent $i$ and agent $j$. The interaction latent $\bm{z}_{ij}$ has $K$ unnormalized dimensions, each indicating a type of interaction. We assume that agents can maintain diverse interactions and that the neighboring agents of the target agent have different importance considering each type of interaction. Consequently, instead of normalizing the dimensions of interaction latent $\bm{z}_{ij}$, we normalize the same dimension of interaction latent between agent $i$ and all its neighbors $j$ as:
\begin{equation}
\bm{z}_{ij}^k = \frac{exp(\bm{z}_{ij}^k)}{\sum_{j{\in}N_i}exp(\bm{z}_{ij}^k)},\quad  \forall k{\in}K.
\end{equation}
$N_i$ denotes the neighbor set of agent $i$, and $K$ represents the number of dimensions of the latent code $\bm{z}_{ij}$, we take the $\bm{z} = \{ z_{ij} \mid 1 \leq i, j \leq N \}$ as the output of the MATE-Encoder.

\begin{table*}[t]
	\fontsize{9}{10}\selectfont
	\resizebox{\textwidth}{!}{
        \setlength\tabcolsep{10pt}
		\renewcommand\arraystretch{1.1}
    		\begin{tabular}{m{5cm}<{\raggedright} || m{1.2cm}<{\centering} | m{1.2cm}<{\centering} | m{1.2cm}<{\centering} | m{1.2cm}<{\centering} | m{1.2cm}<{\centering} | m{1.2cm}<{\centering}}
    		\hline\thickhline
            \rowcolor{mygray}
             & \multicolumn{3}{c|}{PHASE} & \multicolumn{3}{c}{Socialnav} \\
            \cline{2-4}\cline{5-7}
            \rowcolor{mygray}
            \multirow{-2}{*}{Model} & ADE$\downarrow$ & FDE$\downarrow$ & Graph.\(\%\)$\uparrow$ & ADE$\downarrow$ & FDE$\downarrow$ & Graph.\(\%\)$\uparrow$ \\ 
    			\hline\hline
    		 MLP & 1.024 & 1.763 & - & 0.241 & 0.513 & - \\
            GAT-LSTM~\cite{velivckovic2017graph} & 1.545 & 2.527 & 52.94 & 0.306 & 0.527 & 21.83\\
              NRI~\cite{kipf2018NRI} & 0.986 & 1.772 & 55.30 & 0.217 & 0.386 & 57.18\\
            EvolveGraph~\cite{li2020evolvegraph} & 0.848 & 1.522 & 58.96 & 0.160 & 0.321 & 70.23\\
            RFM(skip 1)~\cite{tacchetti2018relational} & 0.870 & 1.581 & 55.30 & 0.160 & 0.325 & 70.05\\ 
            RFM~\cite{tacchetti2018relational}         & 0.892 & 1.630 & 54.71 & 0.156 & 0.317 & 71.53\\ 
            fNRI~\cite{webb2019factorised}        & 0.883 & 1.607 & 55.49 & 0.151 & 0.308 & 33.97\\ 
            IMMA~\cite{sun2022IMMA}        & 0.801 & 1.484 & 79.21 & 0.139 & 0.279 & 81.38\\
           \hline
            \textbf{Ours} & \textbf{0.660} & \textbf{1.281} & \textbf{91.18}& \textbf{0.111} & \textbf{0.229} & \textbf{91.21}\\
			\hline
    		\end{tabular}
	}
    \caption{Quantitative results of trajectory prediction on PHASE and Socialnav datasets.}
	\label{tab:PHASE-socialnav}
    \vspace{-10pt}
\end{table*}

\noindent\textbf{MATE-Decoder with Neural Interaction Energy Module.}
\label{decoder}
To optimize the prediction of the future trajectory, our MATE-Decoder incorporates a sequential model, incorporating our proposed constraints. Specifically in the MATE-Decoder, we introduce a Neural Interaction Energy Module (NIEM) unit (depicted in Fig. \ref{Fig.network}) that processes the previous position $\bm{x}$, velocity $\bm{v}$, interaction latent $\bm{z}$, and previous hidden state $\bm{h}$, yielding the new position, velocity, and hidden state for the predicted step. To enhance the performance of the MATE-Decoder, we adopt a burn-in strategy~\cite{kipf2018NRI} that initiates the decoding process from the first observation step during training, thereby prolonging the prediction horizon and enhancing model robustness. 

\begin{figure}[t]
  \centering
  \includegraphics[width=0.47\textwidth]{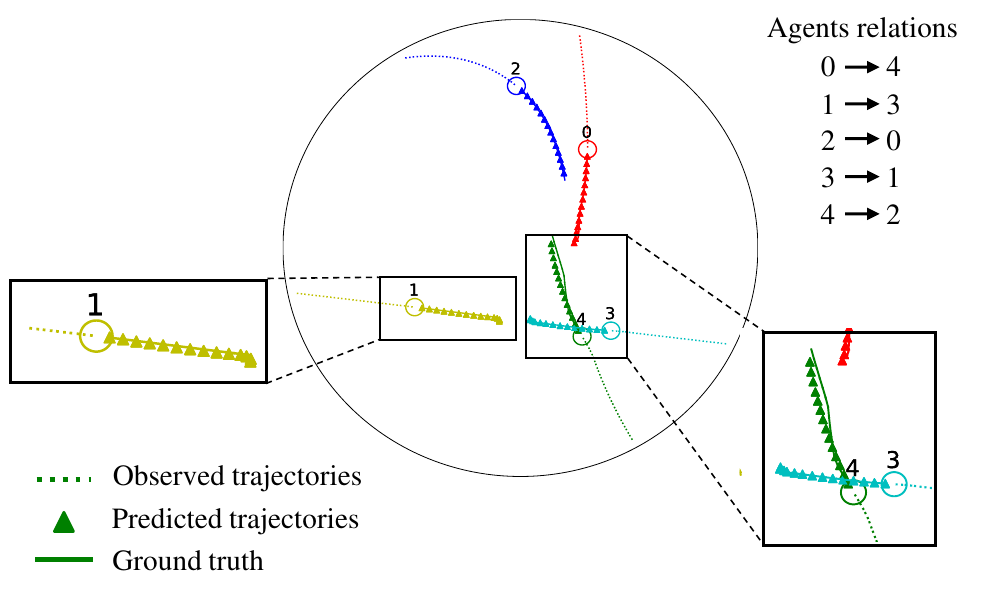}
  \caption{Qualitative analysis of our model on Socialnav dataset. Agent relations mean that the agent to the left of the arrow moves towards the agent to the right of the arrow.}
  \vspace{-10pt}
  \label{Fig.trajectory and heatmap of Socialnav}
\end{figure}


\noindent\textbf{Structure of the NIEM.} Initially, a message passing (MSG) block is leveraged to integrate the information between the hidden state $\bm{h}$ and interaction latent $\bm{z}$. This is followed by the processing of the embedding from the previous step's kinematic states along with the output from the MSG block through a GRU module. The GRU module results in the generation of an updated hidden state. Subsequently, an Energy Module calculates the neural interaction energy features $\bm{E}$ for each agent. These features serve as inputs for subsequent networks, facilitating the generation of agent trajectories. Additionally, a Constraint Module exploits these neural interaction energy features to formulate the inter-agent interaction constraint $\mathcal{L}_{E}$ and the intra-agent motion constraint $\mathcal{L}_{D}$. It is noteworthy that, in our implementation, the inter-agent interaction constraint is integrated into the trajectory prediction through two distinct methods. For each dataset, the method yielding the most favorable result is selected for final result presentation. Refer to the Appendix for more details about the NIEM.

\subsection{Loss for Training}
We employ an end-to-end training method. In our approach, we compute the mean squared error (MSE), denoted as $\mathcal{L}_{P}$, between the predicted position and their corresponding ground truth across period $T$. That is,
\begin{equation}
\begin{aligned}
    \mathcal{L}_{P} &= \frac{1}{N}\frac{1}{T}\sum_{i=1}^{N}\sum_{t=1}^{T}\|\bm{x}_{i}^{t}-\bm{d}_{i}^{t}\|_2,
\end{aligned}
\end{equation}
where $\bm{d}_i^t$ means ground truth position for agent $i$ at time step $t$. Moreover, we integrate the proposed constraints $L_{E}$ and $L_{D}$ into our optimization. Above all, the loss function for our framework is shown below and $\lambda_1$, $\lambda_2$ are hyperparameters for scale balance, 
\begin{equation}
\label{loss}
    \mathcal{L} = \mathcal{L}_{P} + \lambda_1 \mathcal{L}_{E} + \lambda_2  \mathcal{L}_{D}.
\end{equation}


\section{Experiments}
\label{Experiments}
\subsection{Datasets}

We validate our method on four datasets: \textbf{PHASE}~\cite{netanyahu2021phase}, \textbf{Socialnav}~\cite{chen2019crowd}, \textbf{Charged}~\cite{GRIN}, \textbf{NBA}. For simulated datasets, we set the same simulation configurations with previous work for fair Comparison. The detailed description each dataset are as follows:

\noindent\textbf{PHASE.} The PHASE is a dataset of physically grounded abstract social events that resemble a wide range of real-life social interactions by including social concepts such as helping another agent. For comparison with previous work, we choose the ``collaboration'' task in that two agents collaborate to move one of the balls to a pre-set position. Note that we additionally take the label information indicating the difference between agents and balls as part of the input. We do the same data augmentation on the training set as previous work, by flipping the environment and rotating it by 90 degrees, 180 degrees, and 270 degrees, resulting in a training set with 8x more instances. We use 80\% for training, 10\% for validation, and 10\% for testing, and predict 10 steps based on the observation of 24 steps. The timestep of this dataset is 0.25.

\noindent\textbf{Social Navigation Environment.} The Social Navigation Environment (Socialnav) dataset includes annotations for each trajectory, and it also includes annotations for social interactions between agents, such as whether two agents are nearby or whether one agent is following another. Complex unseen trajectories can be easily generated by varying different parameters and configurations. We follow the environmental configurations of previous work. In this simulation, agents are controlled by ORCA policy. We set the radius of agents to 0.3, the radius of the circular environment to 8, and the preferred speed of agents to 1. For Social Navigation we use 100k multi-agent trajectories, in total for training, validation, and testing. We also simulate 50k, 25k, 12.5k, and 10k samples to validate the performance of our model in different dataset sizes. We use 80\% for training, 10\% for validation, and 10\% for testing, and predict 10 steps based on the observation of 24 steps. The timestep of this dataset is 0.25. 

\noindent\textbf{Charged.} The Charged datasets are usually used to illustrate the performance of graph relation prediction in previous work, as the relation type between agents can be set manually. We simulate a dynamic system controlled by physical laws. There are 5 charged particles in each scene, each particle has a positive or negative charge with equal probability, attracting the particles with different charges and vice versa. We experiment with the generalization of models on the Charged dataset by modifying configurations. We generate 50k trajectories for training, 10k trajectories for validation, and 10k trajectories for testing. In the basic environment, we set the side length of the box square to 5, the number of particles to 5, and the sampling timestep to 0.2. To compare with GRIN and previous work, we use our model to predict 20 steps based on the observation of 80 steps. The timestep of this dataset is 0.2.

\noindent\textbf{NBA.} The National Basketball Association (NBA) uses the SportVU tracking system to collect player-tracking data, where each frame contains the location of all ten players and the ball at each time step. We use 300k multi-agent trajectories of players and the ball in total for training, validation, and testing. We use 80\% for training, 10\% for validation, and 10\% for testing, and predict 10 steps based on the observation of 24 steps. The timestep of this dataset is 0.4.

\subsection{Setup}
\textbf{Baselines.} Following IMMA~\cite{sun2022IMMA} and GRIN~\cite{GRIN}, we compare our results with approaches conducting multi-agent trajectory prediction. In specific, we choose MLP, GAT-LSTM~\cite{velivckovic2017graph}, NRI~\cite{kipf2018NRI}, EvolveGraph~\cite{li2020evolvegraph}, RFM~\cite{tacchetti2018relational}, fNRI~\cite{webb2019factorised}, IMMA, FQA~\cite{kamra2020fqa}, dNRI~\cite{graber2020dynamic}, Social-GAN~\cite{gupta2018social}, and GRIN as baselines in quantitative evaluation. We choose GAT-LSTM and IMMA for qualitative analysis.

\noindent\textbf{Hyperparameter Choice.} For all datasets, we set $\lambda_1$ to 1 and $\lambda_2$ to 0.001 in Eq. (\ref{loss}). $\lambda_2$ is set to a smaller value to balance the regularization term. When comparing with baselines, we only report this set of hyperparameters as they have the highest average performance across four datasets.

\noindent\textbf{Evaluation Metrics.} We quantitatively compare our model with other methods using the following metrics:
1) Average Displacement Error (\textbf{ADE}) is the L2 distance between the ground truth trajectories and predicted trajectories. 2) Final Displacement Error (\textbf{FDE}) measures the L2 distance between the ground truth of the destination and the predicted destination. 3) Graph Accuracy  (\textbf{Graph.}) measures the consistency between the predicted interaction graph and the input interaction graph. To ensure a fair comparison with previous methods~\cite{sun2022IMMA}, we use Graph Accuracy to evaluate our model on the PHASE and Socialnav datasets since these datasets have the corresponding ground-truth social interaction graph. 

\noindent\textbf{Implementation Details.}
For generalization in the Socialnav dataset, we change the environment by setting the number of agents to 10 (doubling the number of agents) and setting the preferred speed of agents to 2 (doubling the speed of agents). For generalization in the Charged dataset, we change the environment by setting the number of agents to 10 (doubling the number of agents), setting the time step to 0.1 (halving the sampling timestep), setting the side length of the box square to 3.5 (halving the environment space size).

We do experiments for PHASE, Socialnav, and Charged datasets with Nvidia 2080Ti GPU, with Ubuntu 18.04 system. Our networks were implemented using the PyTorch framework. For Socialnav and Charged datasets, the input dimension is 2 (position for each agent) and it is 4 for the PHASE dataset (position and label for each agent). We train the model with Adam optimizer with learning rate $1 \times 10^{-3}$ and decay the learning rate by a factor of 0.9 if the validation performance has not improved in 5 epochs. We use a hidden size of 96, 128, and 128 for the Social Navigation Environment, PHASE, and the Charged dataset, and we use 4 latent graph layers for PHASE and Socialnav datasets (the dimension of edge latent $\textbf{z}$ is 4) and 4 latent graph layers for the Charged dataset (the dimension of edge latent $\textbf{z}$ is 2). 

As the size of the NBA dataset is much larger than the other three datasets, we do an experiment using Nvidia 3090 GPU, with Adam optimizer with learning rate $5 \times 10^{-4}$ and decay the learning rate by a factor of 0.9 if the validation performance has not improved in 5 epochs. We use a hidden size of 256 and 5 latent graph layers for the NBA dataset (the dimension of edge latent $\textbf{z}$ is 5).

For implementing IMMA, RFM, RFM (skip 1), and GAT models for comparison in the Charged dataset, we use the default configurations mentioned in IMMA, that is, we use Adam optimizer with an initial learning rate of 1e-6 and decay the learning rate by
a factor of 0.9 if the validation performance has not improved in 5 epochs, and we use 2 latent graph layers (the dimension of edge latent $\textbf{z}$ is 2).

\subsection{Results for Trajectory Prediction}

\textbf{Quantitative Evaluation.} In Table \ref{tab:PHASE-socialnav}, we report ADE, FDE, and Graph Accuracy for each model on PHASE and Socialnav datasets. Our model achieves the best performance on both datasets. Compared with the IMMA model, we achieved a relative improvement by 17.6\% $\slash$ 13.7\% in ADE $\slash$ FDE on the PHASE dataset, respectively. On the Socialnav dataset, the improvement is 20.1\% $\slash$ 17.9\% in ADE $\slash$ FDE. In the aspect of the Graph Accuracy metric, we compare the ability of each model to predict social interaction graphs. The results show that our model achieves an accuracy of over 90\% on both datasets. It means that our method can not only accurately predict the future trajectories of each agent, but also clearly express the relationships between agents, which helps explain how agents affect other’s trajectories.
Our model and other models are also tested on Charged dataset and NBA dataset. Results are presented in Table \ref{tab:Charge} and Table \ref{tab:NBA}. As can be seen, our model yields the best performance compared with existing methods, indicating that our method is robust to both simulated and real-world scenarios.

\begin{figure}[t]
  \vspace{-3pt}
  \centering
  \includegraphics[width=0.45\textwidth]{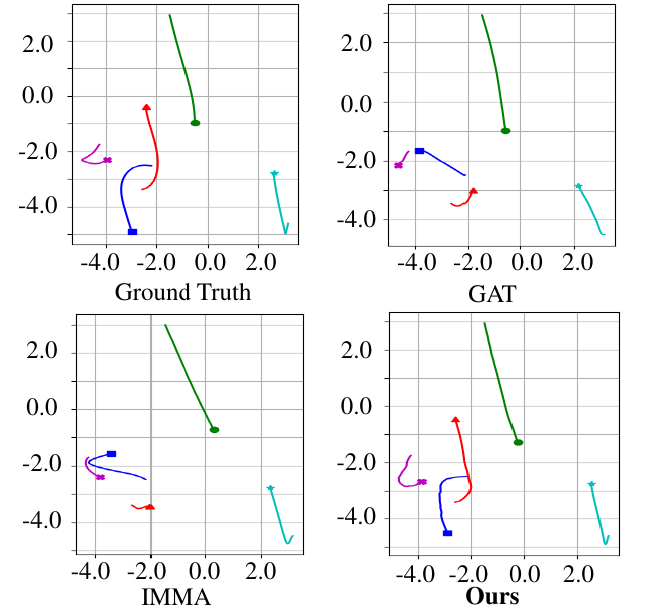}
  \caption{Qualitative analysis of models on the Charged dataset. Different colored lines denote the trajectories of different charges.}
  \vspace{-5pt}
  \label{Fig.trajectory of Particle}
\end{figure}

\begin{figure}[t]
  \centering
  \includegraphics[width=0.46\textwidth]{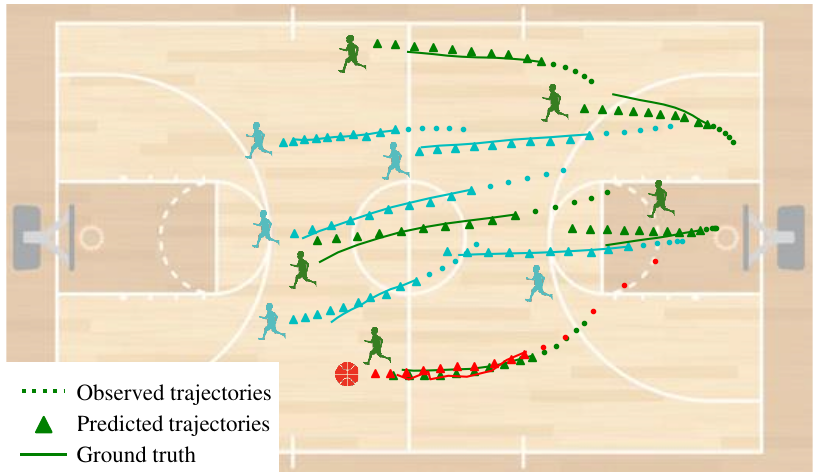}
  \caption{Qualitative analysis of our model on NBA dataset. The red agent is the basketball, and the other colored agents represent players.}
  \vspace{-5pt}
  \label{Fig.trajectory of NBA}
\end{figure}

\noindent\textbf{Qualitative Analysis.} Fig. \ref{Fig.trajectory and heatmap of Socialnav} visualizes the trajectory prediction results of our model on Socialnav dataset. Note that each agent has a target agent and moves towards its target. It can be seen that our model accurately predicts the future trajectory of agents with the assistance of their past information and interactions with others. 
We also visualize the prediction results of our method and others in Fig. \ref{Fig.trajectory of Particle} on Charged dataset. It can be seen that interactions exist between charges that repel and attract each other. We accurately predict such interactions and future trajectories of charges. We visualize the results on NBA dataset in Fig. \ref{Fig.trajectory of NBA} and show the trajectories of the ball and the players in different colored lines. 
See more visualizations of our model in the Appendix.

\begin{table}[t]
    	\centering
        \fontsize{9}{10}\selectfont
        \resizebox{0.48\textwidth}{!}{
            \setlength\tabcolsep{4pt}
		  \renewcommand\arraystretch{1.1}
    		\begin{tabular}{m{5cm}<{\raggedright} || m{1.2cm}<{\centering} | m{1.2cm}<{\centering}}
    		 \hline
            \thickhline
            \rowcolor{mygray}
            Model & ADE$\downarrow$ & FDE$\downarrow$ \\
            \hline\hline
    		 NRI~\cite{kipf2018NRI} & 0.63 & 1.30\\ 
             FQA~\cite{kamra2020fqa} & 0.82 & 1.76 \\
             dNRI~\cite{graber2020dynamic} & 0.94 & 1.93 \\
             GAT-LSTM~\cite{velivckovic2017graph}  & 0.74 & 1.45 \\
            Social-GAN~\cite{gupta2018social}   & 0.66 & 1.25 \\
            RFM(skip 1)~\cite{tacchetti2018relational}  & 0.78 & 1.53 \\ 
            RFM~\cite{tacchetti2018relational}          & 0.79 & 1.55 \\ 
            IMMA~\cite{sun2022IMMA}         & 0.73 & 1.44 \\ 
            GRIN~\cite{GRIN}         & 0.52 & 1.09 \\
            \hline
            \textbf{Ours} & \textbf{0.48} & \textbf{1.05} \\
			\hline
    		\end{tabular}
            }
            
    \caption{Quantitative results on the Charged dataset.}
	\label{tab:Charge}
    \vspace{-10pt}
\end{table}

\begin{table}[t]
    \centering
    \fontsize{9}{10}\selectfont
    \resizebox{0.48\textwidth}{!}{
            \setlength\tabcolsep{4pt}
		  \renewcommand\arraystretch{1.2}
    		\begin{tabular}{m{5cm}<{\raggedright} || m{1.2cm}<{\centering} | m{1.2cm}<{\centering}}
            \hline
            \thickhline
            \rowcolor{mygray}
            Model & ADE$\downarrow$ & FDE$\downarrow$ \\
            \hline\hline
    		 MLP & 1.113 & 1.990\\ 
             GAT-LSTM~\cite{velivckovic2017graph} & 0.978 & 1.733 \\
             NRI~\cite{kipf2018NRI} & 0.946 & 1.818 \\
              EvolveGraph~\cite{li2020evolvegraph}  & 0.896 & 1.695 \\
            RFM(skip 1)~\cite{tacchetti2018relational}   & 0.938 & 1.756 \\
            RFM~\cite{tacchetti2018relational}  & 0.839 & 1.572 \\ 
            fNRI~\cite{webb2019factorised}          & 0.804 & 1.517 \\ 
            IMMA~\cite{sun2022IMMA}         & 0.769 & 1.438 \\
            \hline
            \textbf{Ours} & \textbf{0.659} & \textbf{1.266} \\
			\hline
    		\end{tabular}
    }
    \caption{Quantitative results on NBA dataset.}
	\label{tab:NBA}
    \vspace{-10pt}
\end{table}

\begin{figure}[t]
  \centering
  \includegraphics[width=0.45\textwidth]{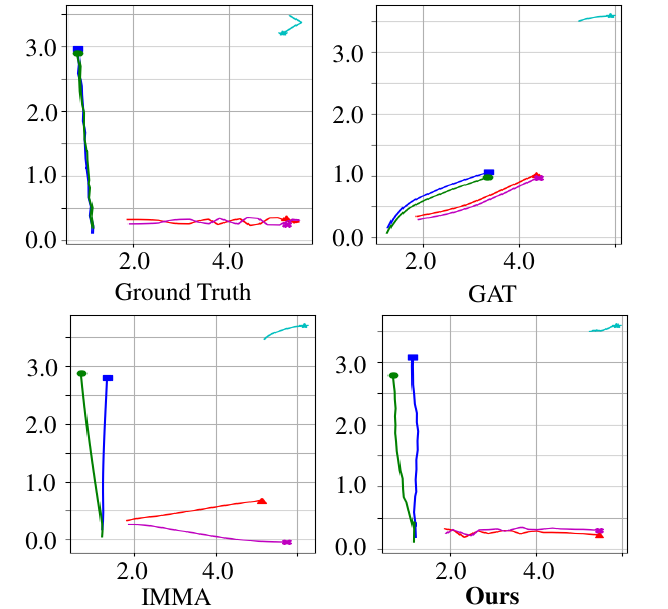}
   \caption{Visualization of generalization on the Charged dataset. Different colored lines denote the trajectories of different charges.}
   \vspace{-10pt}
  \label{Fig.zero-shot of Particle}
\end{figure}

\begin{table}[t]
  \centering
  \fontsize{10}{12}\selectfont
  \resizebox{0.48\textwidth}{!}{
  \setlength\tabcolsep{8pt}
  \renewcommand\arraystretch{1.0}
 \begin{tabular}{m{1.5cm}<{\centering} | m{3cm}<{\centering} || m{1cm}<{\centering} | m{1cm}<{\centering}}
    \hline
    \thickhline 
    \rowcolor{mygray}
    Dataset  &Method     & ADE$\downarrow$    & FDE$\downarrow$    \\
    \hline\hline
    \multirow{4}{*}{PHASE}  &MATE & 0.706  & 1.462       \\
    &MATE + $\mathcal{L}_{E}$  & 0.688  & 1.357       \\
    &MATE + $\mathcal{L}_{D}$  & 0.690  & 1.382       \\
   &MATE + $\mathcal{L}_{E}$ + $\mathcal{L}_{D}$    & \textbf{0.660}  & \textbf{1.281} \\
    \bottomrule
  \end{tabular}
  }
  \caption{Ablation study on PHASE dataset.} 
  \label{tab:ablation}
  \vspace{-10pt}
\end{table}

\begin{table*}
	
	\fontsize{9}{10}\selectfont
	\resizebox{\textwidth}{!}{
        \setlength\tabcolsep{10pt}
		\renewcommand\arraystretch{1.1}
    		\begin{tabular}{m{5cm}<{\raggedright} || m{1.2cm}<{\centering} | m{1.2cm}<{\centering} | m{1.2cm}<{\centering} | m{1.2cm}<{\centering} | m{1.2cm}<{\centering} | m{1.2cm}<{\centering}}
    		\hline\thickhline
            \rowcolor{mygray}
             & \multicolumn{3}{c|}{Double agents} & \multicolumn{3}{c}{Double speed} \\
            \cline{2-4}\cline{5-7}
            \rowcolor{mygray}
            \multirow{-2}{*}{Model} & ADE$\downarrow$ & FDE$\downarrow$ & Graph.\(\%\)$\uparrow$ & ADE$\downarrow$ & FDE$\downarrow$ & Graph.\(\%\)$\uparrow$ \\ 
    			\hline\hline
    		 MLP & - & - & - & 0.303 & 0.632 & - \\
            GAT-LSTM~\cite{velivckovic2017graph} & 1.486 & 2.538 & 19.84 & 0.361 & 0.635 & 22.04\\
            NRI~\cite{kipf2018NRI} & 0.527 & 1.096 & 34.57 & 0.269 & 0.485 & 55.10\\
            EvolveGraph~\cite{li2020evolvegraph} & 0.354 & 0.988 & 50.56 & 0.219 & 0.421 & 67.30\\
            RFM(skip 1)~\cite{tacchetti2018relational} & 0.441 & 0.792 & 49.41 & 0.209 & 0.418 & 68.03\\ 
            RFM~\cite{tacchetti2018relational}         & 0.333 & 0.762 & 51.37 & 0.205 & 0.411 & 69.54\\ 
            fNRI~\cite{webb2019factorised}        & 0.310 & 0.659 & 19.71 & 0.206 & 0.410 & 33.48\\ 
            IMMA~\cite{sun2022IMMA}        & 0.195 & 0.406 & 64.25 & 0.192 & 0.383 & 78.84\\
            \hline
            \textbf{Ours} & \textbf{0.173} & \textbf{0.367} & \textbf{78.34}& \textbf{0.159} & \textbf{0.329} & \textbf{89.02}\\
			\hline
    		\end{tabular}
	}
    \caption{Quantitative results of zero-shot generalization on Socialnav dataset.}
	\label{tab:zero-shot-socialnav}
    \vspace{-10pt}
\end{table*}

\begin{table*}[t]
	
	\fontsize{9}{10}\selectfont
	\resizebox{\textwidth}{!}{
        \setlength\tabcolsep{10pt}
		\renewcommand\arraystretch{1.1}
    		\begin{tabular}{m{5cm}<{\raggedright} || m{1.2cm}<{\centering} | m{1.2cm}<{\centering} | m{1.2cm}<{\centering} | m{1.2cm}<{\centering}| m{1.2cm}<{\centering} | m{1.2cm}<{\centering}}
    		\hline\thickhline
            \rowcolor{mygray}
             & \multicolumn{2}{c|}{Double agents} & \multicolumn{2}{c|}{Half sampling $\Delta t$} & \multicolumn{2}{c}{Half space}\\
            \cline{2-3}\cline{4-5}\cline{6-7}
            \rowcolor{mygray}
            \multirow{-2}{*}{Model} & ADE$\downarrow$ & FDE$\downarrow$  & ADE$\downarrow$ & FDE$\downarrow$ & ADE$\downarrow$ & FDE$\downarrow$  \\ 
    			\hline\hline
            GAT-LSTM~\cite{velivckovic2017graph}   & 1.198 & 2.282  & 0.381 & 0.738 & 1.435 & 2.667\\
            RFM(skip 1)~\cite{tacchetti2018relational}  & 4.538 & 10.44  & 0.406 & 0.822 & 1.380 & \textbf{2.5286} \\ 
            RFM~\cite{tacchetti2018relational}          & 1.897 & 4.019  & 0.414 & 0.834 & 1.378 & 2.5287\\  
            IMMA~\cite{sun2022IMMA}        & 1.140 & 2.227  & 0.402 & 0.777 & 1.400 & 2.5290\\
            \hline
            \textbf{Ours} & \textbf{1.011} & \textbf{2.148} & \textbf{0.348} & \textbf{0.745} & \textbf{1.341} & {2.604}\\
		  \hline
    		\end{tabular}
	}
    \caption{Quantitative results of zero-shot generalization on Charged dataset.}
	\label{tab:zero-shot-particle}
    \vspace{-10pt}
\end{table*}


\begin{figure}[t]
  \centering
  \includegraphics[width=0.4\textwidth]{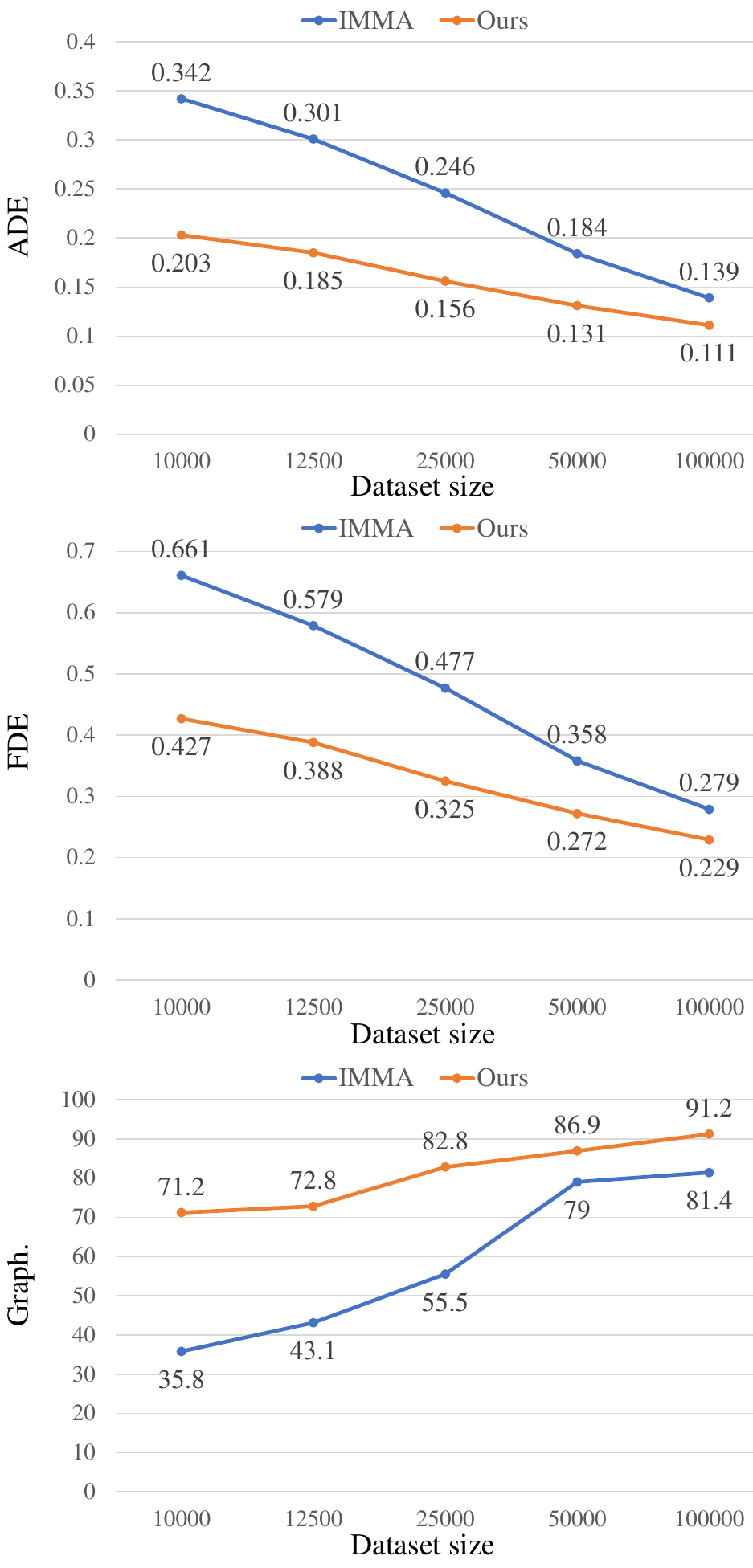}
  \caption{Results of our model and IMMA on different sizes of Socialnav datasets. Lower ADEs and FDEs and higher graph accuracies (Graph.) are better.}
  \label{Fig.dataset}
  \vspace{-10pt}
\end{figure}

\subsection{Ablation Study} Table \ref{tab:ablation} shows the results of our ablation study on PHASE dataset, where we verify the effect of different components of our model. Specifically, we compare the performance of our model with different variants: 1) MATE (base model), 2) MATE + $\mathcal{L}_{E}$ (inter-agent interaction constraint), 3) MATE + $\mathcal{L}_{D}$ (intra-agent motion constraint), and 4) MATE + $\mathcal{L}_{E}$ + $\mathcal{L}_{D}$. We depict the results in Table \ref{tab:ablation}. Compared with MATE (base model), our model performs better with the assistance of the inter-agent interaction constraint on PHASE dataset, indicating that modeling interactions between agents through neural interaction energy and constraining it as prior knowledge preserves a temporally stable system, improving the prediction accuracy and robustness of our model. Moreover, adding the intra-agent motion constraint further boosts the performance, indicating that punishing the incoherent motion prediction and connecting the relation between interactions and motions of agents helps to preserve the agent-level temporal stability.

\subsection{Generalization to Unseen Scenarios}

Our method focuses on preserving temporal stability, which exists in various multi-agent systems. Consequently, our approach is robust to unseen scenarios. We conduct zero-shot experiments to present the generalization ability of our method. 
Table \ref{tab:zero-shot-socialnav} and Table \ref{tab:zero-shot-particle} depict the zero-shot experiments on Socialnav and Charged datasets, where we test the trajectory prediction accuracy of our model and other existing models to unseen situations.

On Socialnav dataset, we create two new environments by 1) doubling the number of agents and 2) doubling the speed of agents. As shown in Table \ref{tab:zero-shot-socialnav}, our model outperforms other methods in both environments, demonstrating that our model is generalizable to different scenarios of complexity and dynamics in the social navigation environment. 

On Charged dataset, we test our model on three novel situations: 1) doubling the number of charges, 2) halving the sampling timestep, and 3) halving the environment space size. As shown in Table \ref{tab:zero-shot-particle}, our model achieves the best performance in most situations. In Fig. \ref{Fig.zero-shot of Particle}, we also visualize the results of our model and others in the unseen situation by halving the environment space size. Though the trajectory prediction accuracy of each model drops, our model performs the best. This indicates that our model has a better generalization ability.

\subsection{Results of Different Dataset Sizes}

To investigate the impact of different size datasets on our models, we simulate datasets of 50k, 25k, 12.5k, and 10k samples, and use the same network structure and hyperparameters, The result is in Fig. \ref{Fig.dataset}. Compared with the strongest baseline (IMMA), our model outperforms on all datasets of different sizes, and the prediction performance remains stable.

\section{Conclusion}

In this work, we proposed a framework named MATE for multi-agent trajectory prediction, aiming at eliminating the prediction fluctuations and preserving the temporal stability of multi-agent systems. Our framework introduces the concept of neural interaction energy to model the interactions and motions of agents in a dynamic environment. Furthermore, we employ partial differential equations to establish the inter-agent interaction constraint and the intra-agent motion constraint, which preserve the temporal stability of the multi-agent system and improve the prediction accuracy of our model. Through quantitative and qualitative experiments, we evaluate our framework on four datasets and demonstrate that it outperforms the state-of-the-art methods in trajectory prediction. We also show the advantages of our framework in zero-shot generalization capacity. Our work opens up new possibilities for representing interactions between agents for multi-agent systems and illustrating the mechanism of how interactions influence agents' future trajectories.
\bibliographystyle{ACM-Reference-Format}
\bibliography{main}


\begin{thebibliography}{49}


\ifx \showCODEN    \undefined \def \showCODEN     #1{\unskip}     \fi
\ifx \showDOI      \undefined \def \showDOI       #1{#1}\fi
\ifx \showISBNx    \undefined \def \showISBNx     #1{\unskip}     \fi
\ifx \showISBNxiii \undefined \def \showISBNxiii  #1{\unskip}     \fi
\ifx \showISSN     \undefined \def \showISSN      #1{\unskip}     \fi
\ifx \showLCCN     \undefined \def \showLCCN      #1{\unskip}     \fi
\ifx \shownote     \undefined \def \shownote      #1{#1}          \fi
\ifx \showarticletitle \undefined \def \showarticletitle #1{#1}   \fi
\ifx \showURL      \undefined \def \showURL       {\relax}        \fi
\providecommand\bibfield[2]{#2}
\providecommand\bibinfo[2]{#2}
\providecommand\natexlab[1]{#1}
\providecommand\showeprint[2][]{arXiv:#2}

\bibitem[Alahi et~al\mbox{.}(2016)]%
        {alahi2016social}
\bibfield{author}{\bibinfo{person}{Alexandre Alahi}, \bibinfo{person}{Kratarth Goel}, \bibinfo{person}{Vignesh Ramanathan}, \bibinfo{person}{Alexandre Robicquet}, \bibinfo{person}{Li Fei-Fei}, {and} \bibinfo{person}{Silvio Savarese}.} \bibinfo{year}{2016}\natexlab{}.
\newblock \showarticletitle{Social lstm: Human trajectory prediction in crowded spaces}. In \bibinfo{booktitle}{\emph{CVPR}}.
\newblock


\bibitem[Cao et~al\mbox{.}(2020)]%
        {cao2020spectral}
\bibfield{author}{\bibinfo{person}{Defu Cao}, \bibinfo{person}{Yujing Wang}, \bibinfo{person}{Juanyong Duan}, \bibinfo{person}{Ce Zhang}, \bibinfo{person}{Xia Zhu}, \bibinfo{person}{Congrui Huang}, \bibinfo{person}{Yunhai Tong}, \bibinfo{person}{Bixiong Xu}, \bibinfo{person}{Jing Bai}, \bibinfo{person}{Jie Tong}, {et~al\mbox{.}}} \bibinfo{year}{2020}\natexlab{}.
\newblock \showarticletitle{Spectral temporal graph neural network for multivariate time-series forecasting}. In \bibinfo{booktitle}{\emph{NeurIPS}}.
\newblock


\bibitem[Chen et~al\mbox{.}(2019)]%
        {chen2019crowd}
\bibfield{author}{\bibinfo{person}{Changan Chen}, \bibinfo{person}{Yuejiang Liu}, \bibinfo{person}{Sven Kreiss}, {and} \bibinfo{person}{Alexandre Alahi}.} \bibinfo{year}{2019}\natexlab{}.
\newblock \showarticletitle{Crowd-robot interaction: Crowd-aware robot navigation with attention-based deep reinforcement learning}. In \bibinfo{booktitle}{\emph{ICRA}}.
\newblock


\bibitem[Chen et~al\mbox{.}(2018)]%
        {chen2018neural}
\bibfield{author}{\bibinfo{person}{Ricky~TQ Chen}, \bibinfo{person}{Yulia Rubanova}, \bibinfo{person}{Jesse Bettencourt}, {and} \bibinfo{person}{David~K Duvenaud}.} \bibinfo{year}{2018}\natexlab{}.
\newblock \showarticletitle{Neural ordinary differential equations}. In \bibinfo{booktitle}{\emph{NeurIPS}}.
\newblock


\bibitem[Chli et~al\mbox{.}(2003)]%
        {chli2003stability}
\bibfield{author}{\bibinfo{person}{Maria Chli}, \bibinfo{person}{Philippe De~Wilde}, \bibinfo{person}{Jan Goossenaerts}, \bibinfo{person}{Vladimir Abramov}, \bibinfo{person}{Nick Szirbik}, \bibinfo{person}{Luis Correia}, \bibinfo{person}{Pedro Mariano}, {and} \bibinfo{person}{Rita Ribeiro}.} \bibinfo{year}{2003}\natexlab{}.
\newblock \showarticletitle{Stability of multi-agent systems}. In \bibinfo{booktitle}{\emph{SMC'03 Conference Proceedings. 2003 IEEE International Conference on Systems, Man and Cybernetics. Conference Theme-System Security and Assurance}}.
\newblock


\bibitem[Deo and Trivedi(2018)]%
        {deo2018convolutional}
\bibfield{author}{\bibinfo{person}{Nachiket Deo} {and} \bibinfo{person}{Mohan~M Trivedi}.} \bibinfo{year}{2018}\natexlab{}.
\newblock \showarticletitle{Convolutional social pooling for vehicle trajectory prediction}. In \bibinfo{booktitle}{\emph{CVPR Workshops}}.
\newblock


\bibitem[Desmond~Zhong et~al\mbox{.}(2020)]%
        {desmond2019symplectic}
\bibfield{author}{\bibinfo{person}{Yaofeng Desmond~Zhong}, \bibinfo{person}{Biswadip Dey}, {and} \bibinfo{person}{Amit Chakraborty}.} \bibinfo{year}{2020}\natexlab{}.
\newblock \showarticletitle{Symplectic ODE-Net: Learning Hamiltonian Dynamics with Control}. In \bibinfo{booktitle}{\emph{ICLR}}.
\newblock


\bibitem[Fassmeyer et~al\mbox{.}(2022)]%
        {fassmeyer2022semi}
\bibfield{author}{\bibinfo{person}{Dennis Fassmeyer}, \bibinfo{person}{Pascal Fassmeyer}, {and} \bibinfo{person}{Ulf Brefeld}.} \bibinfo{year}{2022}\natexlab{}.
\newblock \showarticletitle{Semi-Supervised Generative Models for Multiagent Trajectories}. In \bibinfo{booktitle}{\emph{NeurIPS}}.
\newblock


\bibitem[Girase et~al\mbox{.}(2019)]%
        {girase2021loki}
\bibfield{author}{\bibinfo{person}{Harshayu Girase}, \bibinfo{person}{Haiming Gang}, \bibinfo{person}{Srikanth Malla}, \bibinfo{person}{Jiachen Li}, \bibinfo{person}{Akira Kanehara}, \bibinfo{person}{Karttikeya Mangalam}, {and} \bibinfo{person}{Chiho Choi}.} \bibinfo{year}{2019}\natexlab{}.
\newblock \showarticletitle{Loki: Long term and key intentions for trajectory prediction}. In \bibinfo{booktitle}{\emph{CVPR}}.
\newblock


\bibitem[Giuliari et~al\mbox{.}(2021)]%
        {giuliari2021transformer}
\bibfield{author}{\bibinfo{person}{Francesco Giuliari}, \bibinfo{person}{Irtiza Hasan}, \bibinfo{person}{Marco Cristani}, {and} \bibinfo{person}{Fabio Galasso}.} \bibinfo{year}{2021}\natexlab{}.
\newblock \showarticletitle{Transformer networks for trajectory forecasting}. In \bibinfo{booktitle}{\emph{ICPR}}.
\newblock


\bibitem[Graber and Schwing(2020)]%
        {graber2020dynamic}
\bibfield{author}{\bibinfo{person}{Colin Graber} {and} \bibinfo{person}{Alexander Schwing}.} \bibinfo{year}{2020}\natexlab{}.
\newblock \showarticletitle{Dynamic neural relational inference for forecasting trajectories}. In \bibinfo{booktitle}{\emph{CVPR Workshops}}.
\newblock


\bibitem[Gupta et~al\mbox{.}(2018)]%
        {gupta2018social}
\bibfield{author}{\bibinfo{person}{Agrim Gupta}, \bibinfo{person}{Justin Johnson}, \bibinfo{person}{Li Fei-Fei}, \bibinfo{person}{Silvio Savarese}, {and} \bibinfo{person}{Alexandre Alahi}.} \bibinfo{year}{2018}\natexlab{}.
\newblock \showarticletitle{Social gan: Socially acceptable trajectories with generative adversarial networks}. In \bibinfo{booktitle}{\emph{CVPR}}.
\newblock


\bibitem[Hauri et~al\mbox{.}(2021)]%
        {hauri2021multi}
\bibfield{author}{\bibinfo{person}{Sandro Hauri}, \bibinfo{person}{Nemanja Djuric}, \bibinfo{person}{Vladan Radosavljevic}, {and} \bibinfo{person}{Slobodan Vucetic}.} \bibinfo{year}{2021}\natexlab{}.
\newblock \showarticletitle{Multi-Modal Trajectory Prediction of NBA Players}. In \bibinfo{booktitle}{\emph{CVPR}}.
\newblock


\bibitem[Helbing and Molnar(1995)]%
        {helbing1995social}
\bibfield{author}{\bibinfo{person}{Dirk Helbing} {and} \bibinfo{person}{Peter Molnar}.} \bibinfo{year}{1995}\natexlab{}.
\newblock \showarticletitle{Social force model for pedestrian dynamics}.
\newblock \bibinfo{journal}{\emph{Physical review E}} \bibinfo{volume}{51}, \bibinfo{number}{5} (\bibinfo{year}{1995}), \bibinfo{pages}{4282}.
\newblock


\bibitem[Higgins et~al\mbox{.}(2017)]%
        {higgins2016beta}
\bibfield{author}{\bibinfo{person}{Irina Higgins}, \bibinfo{person}{Loic Matthey}, \bibinfo{person}{Arka Pal}, \bibinfo{person}{Christopher Burgess}, \bibinfo{person}{Xavier Glorot}, \bibinfo{person}{Matthew Botvinick}, \bibinfo{person}{Shakir Mohamed}, {and} \bibinfo{person}{Alexander Lerchner}.} \bibinfo{year}{2017}\natexlab{}.
\newblock \showarticletitle{beta-{VAE}: Learning basic visual concepts with a constrained variational framework}. In \bibinfo{booktitle}{\emph{ICLR}}.
\newblock


\bibitem[Huang et~al\mbox{.}(2019)]%
        {huang2019stgat}
\bibfield{author}{\bibinfo{person}{Yingfan Huang}, \bibinfo{person}{Huikun Bi}, \bibinfo{person}{Zhaoxin Li}, \bibinfo{person}{Tianlu Mao}, {and} \bibinfo{person}{Zhaoqi Wang}.} \bibinfo{year}{2019}\natexlab{}.
\newblock \showarticletitle{STGAT: Modeling spatial-temporal interactions for human trajectory prediction}. In \bibinfo{booktitle}{\emph{ICCV}}.
\newblock


\bibitem[Kamra et~al\mbox{.}(2020)]%
        {kamra2020fqa}
\bibfield{author}{\bibinfo{person}{Nitin Kamra}, \bibinfo{person}{Hao Zhu}, \bibinfo{person}{Dweep~Kumarbhai Trivedi}, \bibinfo{person}{Ming Zhang}, {and} \bibinfo{person}{Yan Liu}.} \bibinfo{year}{2020}\natexlab{}.
\newblock \showarticletitle{Multi-agent trajectory prediction with fuzzy query attention}. In \bibinfo{booktitle}{\emph{NeurIPS}}.
\newblock


\bibitem[Kanda et~al\mbox{.}(2002)]%
        {kanda2002development}
\bibfield{author}{\bibinfo{person}{Takayuki Kanda}, \bibinfo{person}{Hiroshi Ishiguro}, \bibinfo{person}{Tetsuo Ono}, \bibinfo{person}{Michita Imai}, {and} \bibinfo{person}{Ryohei Nakatsu}.} \bibinfo{year}{2002}\natexlab{}.
\newblock \showarticletitle{Development and evaluation of an interactive humanoid robot" Robovie"}. In \bibinfo{booktitle}{\emph{ICRA}}.
\newblock


\bibitem[Karamouzas et~al\mbox{.}(2014)]%
        {karamouzas2014universal}
\bibfield{author}{\bibinfo{person}{Ioannis Karamouzas}, \bibinfo{person}{Brian Skinner}, {and} \bibinfo{person}{Stephen~J Guy}.} \bibinfo{year}{2014}\natexlab{}.
\newblock \showarticletitle{Universal power law governing pedestrian interactions}.
\newblock \bibinfo{journal}{\emph{Physical review letters}} \bibinfo{volume}{113}, \bibinfo{number}{23} (\bibinfo{year}{2014}), \bibinfo{pages}{238701}.
\newblock


\bibitem[Karniadakis et~al\mbox{.}(2021)]%
        {karniadakis2021physics}
\bibfield{author}{\bibinfo{person}{George~Em Karniadakis}, \bibinfo{person}{Ioannis~G Kevrekidis}, \bibinfo{person}{Lu Lu}, \bibinfo{person}{Paris Perdikaris}, \bibinfo{person}{Sifan Wang}, {and} \bibinfo{person}{Liu Yang}.} \bibinfo{year}{2021}\natexlab{}.
\newblock \showarticletitle{Physics-informed machine learning}.
\newblock \bibinfo{journal}{\emph{Nature Reviews Physics}} \bibinfo{volume}{3}, \bibinfo{number}{6} (\bibinfo{year}{2021}), \bibinfo{pages}{422--440}.
\newblock


\bibitem[Kingma and Welling(2014)]%
        {kingma2013auto}
\bibfield{author}{\bibinfo{person}{Diederik~P Kingma} {and} \bibinfo{person}{Max Welling}.} \bibinfo{year}{2014}\natexlab{}.
\newblock \showarticletitle{Auto-encoding variational bayes}. In \bibinfo{booktitle}{\emph{ICLR}}.
\newblock


\bibitem[Kipf et~al\mbox{.}(2018)]%
        {kipf2018NRI}
\bibfield{author}{\bibinfo{person}{Thomas~N. Kipf}, \bibinfo{person}{Ethan Fetaya}, \bibinfo{person}{Kuan-Chieh Wang}, \bibinfo{person}{Max Welling}, {and} \bibinfo{person}{Richard~S. Zemel}.} \bibinfo{year}{2018}\natexlab{}.
\newblock \showarticletitle{Neural Relational Inference for Interacting Systems}. In \bibinfo{booktitle}{\emph{ICML}}.
\newblock


\bibitem[Levinson et~al\mbox{.}(2011)]%
        {levinson2011towards}
\bibfield{author}{\bibinfo{person}{Jesse Levinson}, \bibinfo{person}{Jake Askeland}, \bibinfo{person}{Jan Becker}, \bibinfo{person}{Jennifer Dolson}, \bibinfo{person}{David Held}, \bibinfo{person}{Soeren Kammel}, \bibinfo{person}{J~Zico Kolter}, \bibinfo{person}{Dirk Langer}, \bibinfo{person}{Oliver Pink}, \bibinfo{person}{Vaughan Pratt}, {et~al\mbox{.}}} \bibinfo{year}{2011}\natexlab{}.
\newblock \showarticletitle{Towards fully autonomous driving: Systems and algorithms}. In \bibinfo{booktitle}{\emph{IV}}.
\newblock


\bibitem[Li et~al\mbox{.}(2020)]%
        {li2020evolvegraph}
\bibfield{author}{\bibinfo{person}{Jiachen Li}, \bibinfo{person}{Fan Yang}, \bibinfo{person}{Masayoshi Tomizuka}, {and} \bibinfo{person}{Chiho Choi}.} \bibinfo{year}{2020}\natexlab{}.
\newblock \showarticletitle{Evolvegraph: Multi-agent trajectory prediction with dynamic relational reasoning}. In \bibinfo{booktitle}{\emph{NeurIPS}}.
\newblock


\bibitem[Li et~al\mbox{.}(2021a)]%
        {GRIN}
\bibfield{author}{\bibinfo{person}{Longyuan Li}, \bibinfo{person}{Jian Yao}, \bibinfo{person}{Li Wenliang}, \bibinfo{person}{Tong He}, \bibinfo{person}{Tianjun Xiao}, \bibinfo{person}{Junchi Yan}, \bibinfo{person}{David Wipf}, {and} \bibinfo{person}{Zheng Zhang}.} \bibinfo{year}{2021}\natexlab{a}.
\newblock \showarticletitle{GRIN Generative Relation and Intention Network for Multi-agent Trajectory Prediction}. In \bibinfo{booktitle}{\emph{NeurIPS}}.
\newblock


\bibitem[Li et~al\mbox{.}(2021b)]%
        {li2021physics}
\bibfield{author}{\bibinfo{person}{Zongyi Li}, \bibinfo{person}{Hongkai Zheng}, \bibinfo{person}{Nikola Kovachki}, \bibinfo{person}{David Jin}, \bibinfo{person}{Haoxuan Chen}, \bibinfo{person}{Burigede Liu}, \bibinfo{person}{Kamyar Azizzadenesheli}, {and} \bibinfo{person}{Anima Anandkumar}.} \bibinfo{year}{2021}\natexlab{b}.
\newblock \showarticletitle{Physics-informed neural operator for learning partial differential equations}.
\newblock \bibinfo{journal}{\emph{arXiv preprint arXiv:2111.03794}} (\bibinfo{year}{2021}).
\newblock


\bibitem[Manh and Alaghband(2018)]%
        {manh2018scene}
\bibfield{author}{\bibinfo{person}{Huynh Manh} {and} \bibinfo{person}{Gita Alaghband}.} \bibinfo{year}{2018}\natexlab{}.
\newblock \showarticletitle{Scene-lstm: A model for human trajectory prediction}.
\newblock \bibinfo{journal}{\emph{arXiv preprint arXiv:1808.04018}} (\bibinfo{year}{2018}).
\newblock


\bibitem[Mao et~al\mbox{.}(2023)]%
        {mao2023leapfrog}
\bibfield{author}{\bibinfo{person}{Weibo Mao}, \bibinfo{person}{Chenxin Xu}, \bibinfo{person}{Qi Zhu}, \bibinfo{person}{Siheng Chen}, {and} \bibinfo{person}{Yanfeng Wang}.} \bibinfo{year}{2023}\natexlab{}.
\newblock \showarticletitle{Leapfrog Diffusion Model for Stochastic Trajectory Prediction}. In \bibinfo{booktitle}{\emph{CVPR}}.
\newblock


\bibitem[Mohamed et~al\mbox{.}(2020)]%
        {mohamed2020social}
\bibfield{author}{\bibinfo{person}{Abduallah Mohamed}, \bibinfo{person}{Kun Qian}, \bibinfo{person}{Mohamed Elhoseiny}, {and} \bibinfo{person}{Christian Claudel}.} \bibinfo{year}{2020}\natexlab{}.
\newblock \showarticletitle{Social-stgcnn: A social spatio-temporal graph convolutional neural network for human trajectory prediction}. In \bibinfo{booktitle}{\emph{CVPR}}.
\newblock


\bibitem[Netanyahu et~al\mbox{.}(2021)]%
        {netanyahu2021phase}
\bibfield{author}{\bibinfo{person}{Aviv Netanyahu}, \bibinfo{person}{Tianmin Shu}, \bibinfo{person}{Boris Katz}, \bibinfo{person}{Andrei Barbu}, {and} \bibinfo{person}{Joshua~B Tenenbaum}.} \bibinfo{year}{2021}\natexlab{}.
\newblock \showarticletitle{PHASE: PHysically-grounded Abstract Social Events for Machine Social Perception}. In \bibinfo{booktitle}{\emph{AAAI}}.
\newblock


\bibitem[Paszke et~al\mbox{.}(2017)]%
        {paszke2017automatic}
\bibfield{author}{\bibinfo{person}{Adam Paszke}, \bibinfo{person}{Sam Gross}, \bibinfo{person}{Soumith Chintala}, \bibinfo{person}{Gregory Chanan}, \bibinfo{person}{Edward Yang}, \bibinfo{person}{Zachary DeVito}, \bibinfo{person}{Zeming Lin}, \bibinfo{person}{Alban Desmaison}, \bibinfo{person}{Luca Antiga}, {and} \bibinfo{person}{Adam Lerer}.} \bibinfo{year}{2017}\natexlab{}.
\newblock \showarticletitle{Automatic differentiation in pytorch}. In \bibinfo{booktitle}{\emph{NeurIPS}}.
\newblock


\bibitem[Pettr{\'e} et~al\mbox{.}(2009)]%
        {pettre2009experiment}
\bibfield{author}{\bibinfo{person}{Julien Pettr{\'e}}, \bibinfo{person}{Jan Ond{\v{r}}ej}, \bibinfo{person}{Anne-H{\'e}l{\`e}ne Olivier}, \bibinfo{person}{Armel Cretual}, {and} \bibinfo{person}{St{\'e}phane Donikian}.} \bibinfo{year}{2009}\natexlab{}.
\newblock \showarticletitle{Experiment-based modeling, simulation and validation of interactions between virtual walkers}. In \bibinfo{booktitle}{\emph{Proceedings of the 2009 ACM SIGGRAPH/Eurographics symposium on computer animation}}.
\newblock


\bibitem[Quan et~al\mbox{.}(2021)]%
        {quan2021holistic}
\bibfield{author}{\bibinfo{person}{Ruijie Quan}, \bibinfo{person}{Linchao Zhu}, \bibinfo{person}{Yu Wu}, {and} \bibinfo{person}{Yi Yang}.} \bibinfo{year}{2021}\natexlab{}.
\newblock \showarticletitle{Holistic LSTM for pedestrian trajectory prediction}.
\newblock \bibinfo{journal}{\emph{IEEE transactions on image processing}}  \bibinfo{volume}{30} (\bibinfo{year}{2021}), \bibinfo{pages}{3229--3239}.
\newblock


\bibitem[Rabinowitz et~al\mbox{.}(2018)]%
        {rabinowitz2018machine}
\bibfield{author}{\bibinfo{person}{Neil Rabinowitz}, \bibinfo{person}{Frank Perbet}, \bibinfo{person}{Francis Song}, \bibinfo{person}{Chiyuan Zhang}, \bibinfo{person}{SM~Ali Eslami}, {and} \bibinfo{person}{Matthew Botvinick}.} \bibinfo{year}{2018}\natexlab{}.
\newblock \showarticletitle{Machine theory of mind}. In \bibinfo{booktitle}{\emph{ICML}}.
\newblock


\bibitem[Raissi et~al\mbox{.}(2019)]%
        {raissi2019physics}
\bibfield{author}{\bibinfo{person}{Maziar Raissi}, \bibinfo{person}{Paris Perdikaris}, {and} \bibinfo{person}{George~E Karniadakis}.} \bibinfo{year}{2019}\natexlab{}.
\newblock \showarticletitle{Physics-informed neural networks: A deep learning framework for solving forward and inverse problems involving nonlinear partial differential equations}.
\newblock \bibinfo{journal}{\emph{Journal of Computational physics}}  \bibinfo{volume}{378} (\bibinfo{year}{2019}), \bibinfo{pages}{686--707}.
\newblock


\bibitem[Sanchez-Gonzalez et~al\mbox{.}(2020)]%
        {sanchez2020learning}
\bibfield{author}{\bibinfo{person}{Alvaro Sanchez-Gonzalez}, \bibinfo{person}{Jonathan Godwin}, \bibinfo{person}{Tobias Pfaff}, \bibinfo{person}{Rex Ying}, \bibinfo{person}{Jure Leskovec}, {and} \bibinfo{person}{Peter Battaglia}.} \bibinfo{year}{2020}\natexlab{}.
\newblock \showarticletitle{Learning to simulate complex physics with graph networks}. In \bibinfo{booktitle}{\emph{ICML}}.
\newblock


\bibitem[Sankaran et~al\mbox{.}(2022)]%
        {sankaranimpact}
\bibfield{author}{\bibinfo{person}{Shyam Sankaran}, \bibinfo{person}{Hanwen Wang}, \bibinfo{person}{Leonardo~Ferreira Guilhoto}, {and} \bibinfo{person}{Paris Perdikaris}.} \bibinfo{year}{2022}\natexlab{}.
\newblock \showarticletitle{On the impact of larger batch size in the training of Physics Informed Neural Networks}. In \bibinfo{booktitle}{\emph{NeurIPS Workshop}}.
\newblock


\bibitem[Sohn et~al\mbox{.}(2015)]%
        {sohn2015learning}
\bibfield{author}{\bibinfo{person}{Kihyuk Sohn}, \bibinfo{person}{Honglak Lee}, {and} \bibinfo{person}{Xinchen Yan}.} \bibinfo{year}{2015}\natexlab{}.
\newblock \showarticletitle{Learning structured output representation using deep conditional generative models}. In \bibinfo{booktitle}{\emph{NeurIPS}}.
\newblock


\bibitem[Sun et~al\mbox{.}(2022)]%
        {sun2022IMMA}
\bibfield{author}{\bibinfo{person}{Fan-Yun Sun}, \bibinfo{person}{Isaac Kauvar}, \bibinfo{person}{Ruohan Zhang}, \bibinfo{person}{Jiachen Li}, \bibinfo{person}{Mykel~J Kochenderfer}, \bibinfo{person}{Jiajun Wu}, {and} \bibinfo{person}{Nick Haber}.} \bibinfo{year}{2022}\natexlab{}.
\newblock \showarticletitle{Interaction modeling with multiplex attention}. In \bibinfo{booktitle}{\emph{NeurIPS}}.
\newblock


\bibitem[Tacchetti et~al\mbox{.}(2019)]%
        {tacchetti2018relational}
\bibfield{author}{\bibinfo{person}{Andrea Tacchetti}, \bibinfo{person}{H.~Francis Song}, \bibinfo{person}{Pedro A.~M. Mediano}, \bibinfo{person}{Vinicius Zambaldi}, \bibinfo{person}{János Kramár}, \bibinfo{person}{Neil~C. Rabinowitz}, \bibinfo{person}{Thore Graepel}, \bibinfo{person}{Matthew Botvinick}, {and} \bibinfo{person}{Peter~W. Battaglia}.} \bibinfo{year}{2019}\natexlab{}.
\newblock \showarticletitle{Relational Forward Models for Multi-Agent Learning}. In \bibinfo{booktitle}{\emph{ICLR}}.
\newblock


\bibitem[Tang et~al\mbox{.}(2023)]%
        {tang2023collaborative}
\bibfield{author}{\bibinfo{person}{Bohan Tang}, \bibinfo{person}{Yiqi Zhong}, \bibinfo{person}{Chenxin Xu}, \bibinfo{person}{Wei-Tao Wu}, \bibinfo{person}{Ulrich Neumann}, \bibinfo{person}{Ya Zhang}, \bibinfo{person}{Siheng Chen}, {and} \bibinfo{person}{Yanfeng Wang}.} \bibinfo{year}{2023}\natexlab{}.
\newblock \showarticletitle{Collaborative uncertainty benefits multi-agent multi-modal trajectory forecasting}.
\newblock \bibinfo{journal}{\emph{IEEE Transactions on Pattern Analysis and Machine Intelligence}} (\bibinfo{year}{2023}).
\newblock


\bibitem[Van Den~Berg et~al\mbox{.}(2011)]%
        {van2011reciprocal}
\bibfield{author}{\bibinfo{person}{Jur Van Den~Berg}, \bibinfo{person}{Stephen~J Guy}, \bibinfo{person}{Ming Lin}, {and} \bibinfo{person}{Dinesh Manocha}.} \bibinfo{year}{2011}\natexlab{}.
\newblock \showarticletitle{Reciprocal n-body collision avoidance}. In \bibinfo{booktitle}{\emph{Robotics Research: The 14th International Symposium ISRR}}.
\newblock


\bibitem[Veli{\v{c}}kovi{\'c} et~al\mbox{.}(2018)]%
        {velivckovic2017graph}
\bibfield{author}{\bibinfo{person}{Petar Veli{\v{c}}kovi{\'c}}, \bibinfo{person}{Guillem Cucurull}, \bibinfo{person}{Arantxa Casanova}, \bibinfo{person}{Adriana Romero}, \bibinfo{person}{Pietro Lio}, {and} \bibinfo{person}{Yoshua Bengio}.} \bibinfo{year}{2018}\natexlab{}.
\newblock \showarticletitle{Graph attention networks}. In \bibinfo{booktitle}{\emph{ICLR}}.
\newblock


\bibitem[Vemula et~al\mbox{.}(2018)]%
        {vemula2018social}
\bibfield{author}{\bibinfo{person}{Anirudh Vemula}, \bibinfo{person}{Katharina Muelling}, {and} \bibinfo{person}{Jean Oh}.} \bibinfo{year}{2018}\natexlab{}.
\newblock \showarticletitle{Social attention: Modeling attention in human crowds}. In \bibinfo{booktitle}{\emph{ICRA}}.
\newblock


\bibitem[Webb et~al\mbox{.}(2019)]%
        {webb2019factorised}
\bibfield{author}{\bibinfo{person}{Ezra Webb}, \bibinfo{person}{Ben Day}, \bibinfo{person}{Helena Andres-Terre}, {and} \bibinfo{person}{Pietro Li{\'o}}.} \bibinfo{year}{2019}\natexlab{}.
\newblock \showarticletitle{Factorised neural relational inference for multi-interaction systems}. In \bibinfo{booktitle}{\emph{ICML Workshops}}.
\newblock


\bibitem[Yue et~al\mbox{.}(2022)]%
        {yue2022human}
\bibfield{author}{\bibinfo{person}{Jiangbei Yue}, \bibinfo{person}{Dinesh Manocha}, {and} \bibinfo{person}{He Wang}.} \bibinfo{year}{2022}\natexlab{}.
\newblock \showarticletitle{Human trajectory prediction via neural social physics}. In \bibinfo{booktitle}{\emph{ECCV}}.
\newblock


\bibitem[Zhan et~al\mbox{.}(2019)]%
        {zhan2018generating}
\bibfield{author}{\bibinfo{person}{Eric Zhan}, \bibinfo{person}{Stephan Zheng}, \bibinfo{person}{Yisong Yue}, \bibinfo{person}{Long Sha}, {and} \bibinfo{person}{Patrick Lucey}.} \bibinfo{year}{2019}\natexlab{}.
\newblock \showarticletitle{Generating multi-agent trajectories using programmatic weak supervision}. In \bibinfo{booktitle}{\emph{ICLR}}.
\newblock


\bibitem[Zhao et~al\mbox{.}({[n.\,d.]})]%
        {zhao2019multi}
\bibfield{author}{\bibinfo{person}{Tianyang Zhao}, \bibinfo{person}{Yifei Xu}, \bibinfo{person}{Mathew Monfort}, \bibinfo{person}{Wongun Choi}, \bibinfo{person}{Chris Baker}, \bibinfo{person}{Yibiao Zhao}, \bibinfo{person}{Yizhou Wang}, {and} \bibinfo{person}{Ying~Nian Wu}.} \bibinfo{year}{[n.\,d.]}\natexlab{}.
\newblock \showarticletitle{Multi-agent tensor fusion for contextual trajectory prediction}. In \bibinfo{booktitle}{\emph{CVPR}}.
\newblock


\bibitem[Zubov et~al\mbox{.}(2021)]%
        {zubov2021neuralpde}
\bibfield{author}{\bibinfo{person}{Kirill Zubov}, \bibinfo{person}{Zoe McCarthy}, \bibinfo{person}{Yingbo Ma}, \bibinfo{person}{Francesco Calisto}, \bibinfo{person}{Valerio Pagliarino}, \bibinfo{person}{Simone Azeglio}, \bibinfo{person}{Luca Bottero}, \bibinfo{person}{Emmanuel Luj{\'a}n}, \bibinfo{person}{Valentin Sulzer}, \bibinfo{person}{Ashutosh Bharambe}, {et~al\mbox{.}}} \bibinfo{year}{2021}\natexlab{}.
\newblock \showarticletitle{Neuralpde: Automating physics-informed neural networks (pinns) with error approximations}.
\newblock \bibinfo{journal}{\emph{arXiv preprint arXiv:2107.09443}} (\bibinfo{year}{2021}).
\newblock


\end{thebibliography}

\end{document}